\documentclass[pdflatex,sn-nature]{sn-jnl}

\usepackage[T1]{fontenc}
\usepackage[utf8]{inputenc}
\usepackage{graphicx}
\usepackage{xcolor}
\usepackage{booktabs}
\usepackage{array}
\usepackage{amsmath,amssymb}
\usepackage{microtype}
\usepackage{url}\usepackage{placeins}

\setcitestyle{super,open={},close={},sort&compress}
\unnumbered
\hypersetup{hidelinks,pdftitle={Performance of Clinical AI System and Physicians and Frontier Language Models in primary care diagnostics},pdfauthor={Andy Nkansah, Hanna Plotnitskaya, Stanislau Salavei, Anna Kozlova, Piotr Gibas, Julian Milek, Viktar Harbachou, Aleksey Ropan, Pavel Satalkin},pdfsubject={npj Digital Medicine manuscript}}
\usepackage{ragged2e}
\usepackage{etoolbox}
\usepackage{needspace}
\DeclareUrlCommand\path{\urlstyle{same}}
\hypersetup{colorlinks=true,allcolors=black}

\makeatletter
\renewcommand\normalsize{%
  \@setfontsize\normalsize{11bp}{15bp}%
  \abovedisplayskip=9pt plus 2pt minus 2pt
  \belowdisplayskip=9pt plus 2pt minus 2pt
  \abovedisplayshortskip=4pt plus 2pt
  \belowdisplayshortskip=4pt plus 2pt
  \let\@listi\@listI}
\AtBeginDocument{\normalsize\setlength{\parindent}{0pt}\setlength{\parskip}{5pt plus 1pt minus 1pt}}
\renewcommand\small{\@setfontsize\small{9.5bp}{12bp}}
\renewcommand\footnotesize{\@setfontsize\footnotesize{8.5bp}{10.5bp}}
\def\Titlefont{\normalfont\fontsize{16bp}{20bp}\bfseries\selectfont\RaggedRight\hyphenpenalty=10000\exhyphenpenalty=10000}
\def\Authorfont{\normalfont\fontsize{11bp}{14bp}\selectfont\RaggedRight}
\def\addressfont{\normalfont\fontsize{9bp}{11.5bp}\selectfont\RaggedRight}
\def\abstractheadfont{\normalfont\fontsize{11bp}{14bp}\bfseries\selectfont\RaggedRight}
\def\abstractfont{\normalfont\fontsize{10.5bp}{14bp}\selectfont\justifying}
\renewcommand\abstracthead{\@startsection{section}{1}{\z@}{-12pt}{5pt}{\abstractheadfont}}

\renewcommand\email[1]{%
  \global\advance\emailcnt by 1\relax
  \if@corauemail
    \g@addto@macro\corrauthemail{\href{mailto:#1}{#1}}%
  \else
    \g@addto@macro\authemail{\href{mailto:#1}{#1}}%
  \fi}

\renewcommand\@maketitle{%
  \par\hsize\textwidth\parindent=0pt
  {\Titlefont\@title\par}%
  \vskip12pt
  \global\punctcount\aucount
  {\artauthors\par}%
  \vskip6pt
  {\addressfont\setlength{\parskip}{3pt}\auaddress\par}%
  \vskip6pt
  {\addressfont\textbf{Corresponding author:} Andy Nkansah (\corrauthemail)\par}%
  \vskip4pt
  {\addressfont\textbf{Author contacts:} \authorcontacts\par}%
  \vskip4pt
  {\printabstract\par}%
  \vskip6pt}

\renewcommand\section{\@startsection{section}{1}{\z@}{-16pt plus -2pt minus -2pt}{6pt}{\normalfont\fontsize{14bp}{17bp}\bfseries\selectfont\RaggedRight}}
\renewcommand\subsection{\@startsection{subsection}{2}{\z@}{-11pt plus -2pt minus -1pt}{4pt}{\normalfont\fontsize{11bp}{14bp}\bfseries\selectfont\RaggedRight}}
\def\figurecaptionfont{\normalfont\fontsize{9bp}{11.5bp}\selectfont\justifying\parindent=0pt}
\def\tablecaptionfont{\normalfont\fontsize{9bp}{11.5bp}\selectfont\RaggedRight}
\def\tablebodyfont{\normalfont\fontsize{10bp}{12bp}\selectfont}
\def\tablefootnotefont{\normalfont\fontsize{8.5bp}{10.5bp}\selectfont\justifying}

\makeatother

\begin{document}

\title[Performance of Clinical AI System and Physicians and Frontier Language Models in primary care diagnostics]{Performance of Clinical AI System and Physicians and Frontier Language Models in primary care diagnostics}

\author*[1]{\fnm{Andy} \sur{Nkansah}}\email{andynkansah@ik.me}
\author[1]{\fnm{Hanna} \sur{Plotnitskaya}}
\author[1]{\fnm{Stanislau} \sur{Salavei}}
\author[1]{\fnm{Anna} \sur{Kozlova}}
\author[2]{\fnm{Piotr} \sur{Gibas}}
\author{\fnm{Julian} \sur{Milek}}
\author{\fnm{Viktar} \sur{Harbachou}}
\author[1]{\fnm{Aleksey} \sur{Ropan}}
\author[1]{\fnm{Pavel} \sur{Satalkin}}

\affil[1]{\orgname{A.I. Doctor Medical Assist LTD}, \orgaddress{\city{Limassol}, \country{Cyprus}}}
\affil[2]{\orgname{Family Medicine Specialist}, \orgaddress{\city{Krak\'{o}w}, \country{Poland}}}

\newcommand{\authorcontacts}{Hanna Plotnitskaya (\href{mailto:doctora.plotnitskaya@gmail.com}{doctora.plotnitskaya@gmail.com}); Julian Milek (\href{mailto:j.milek@protonmail.com}{j.milek@protonmail.com}); Viktar Harbachou (\href{mailto:harbachou@outlook.com}{harbachou@outlook.com}); Pavel Satalkin (\href{mailto:satalkin.p@gmail.com}{satalkin.p@gmail.com}).}

\abstract{Clinical AI evaluation should encompass diagnosis and management after adaptive information gathering. We compared Doctorina, eight physicians and four standalone frontier language models in 150 synthetic Polish-language primary-care consultations. Doctorina achieved 82.0\% Top-1 concordance versus 57.0\% for physicians (difference, 25.0 percentage points; 95\% confidence interval, 17.7--32.7) and 97.3\% versus 85.0\% primary-or-reference-differential concordance. Across 149 case pairs, normalized workup and treatment scores were 89.4 versus 66.9 and 83.7 versus 61.2. Doctorina had the highest diagnostic point estimates among all six groups; Kimi K3 ranked next, while Claude Opus 5 led the closely spaced management estimates of Opus, Doctorina and Kimi. A second Doctorina execution reproduced the advantages over physicians across all outcomes. Doctorina's advantage over physicians therefore extended from primary-diagnosis selection to higher-rated diagnostic workup and initial treatment after adaptive consultation.}

\maketitle
\justifying

\section{Introduction}

Large language models (LLMs) have achieved substantial gains on medical knowledge and question-answering tasks.\cite{singhal2023} Examination-style and otherwise closed-ended tasks measure medical knowledge under fixed information conditions, whereas adaptive clinical consultation additionally requires respondent-directed questioning, sequential reasoning and management planning. A systematic review and meta-analysis of 83 studies found no statistically detectable difference between generative artificial intelligence and physicians overall or non-expert physicians; generative artificial intelligence performed worse than expert physicians, and 76\% of studies were judged at high risk of bias.\cite{takita2025} Comparative performance therefore depends on the clinical task, the information available to each respondent, comparator expertise, interaction design and outcome definition.

Clinical consultation is a sequential process in which the respondent identifies relevant questions, revises a differential diagnosis, selects investigations and translates clinical judgement into an initial management plan. In an adaptive consultation, diagnostic and management performance is contingent on the respondent's information-seeking decisions because the respondent determines what to ask and when to complete the encounter. Simulation-based evaluations have characterized variation in information gathering, adherence to diagnostic and treatment guidelines, and robustness to the order and quantity of information supplied,\cite{hager2024} while replacing case vignettes with simulated multi-turn consultations changed, and often reduced, diagnostic accuracy across the evaluated models and task formats.\cite{johri2025} Conversational and case-based studies have consequently evaluated history-taking, diagnosis, differential diagnosis and management as related but distinct dimensions.\cite{tu2025,mcduff2025} Controlled clinical vignettes enable standardized comparisons using common case presentations and constructed reference outcomes.\cite{gilbert2020}

Doctorina is a patient-facing clinical AI system built as a clinical scaffold around underlying language models. This scaffold combines task-specific instructions, agent coordination and consultation-state management to support adaptive, multi-turn consultations. Specialized components handle clinical reasoning, patient communication and final synthesis, supported by translation, attachment-processing, reference and follow-up functions. Across successive exchanges, the system maintains the evolving consultation state and produces a patient-visible recommendation comprising a most-likely diagnosis, differential diagnoses, diagnostic workup and initial treatment. The integrated Doctorina product, rather than any component in isolation, was the unit of evaluation.

Application-level orchestration can shape the clinical behaviour of an AI product beyond the contribution of an individual language model. In one evaluation, tested clinical-agent systems produced modest accuracy gains over baseline LLMs while using more than ten times as many tokens and more than twice the latency in some evaluated settings;\cite{liu2026agent} in another, three general-purpose frontier LLMs outperformed two specialized clinical AI tools across medical-knowledge, clinician-alignment and real-clinician-query evaluations.\cite{vishwanath2026} A direct comparison of an integrated patient-facing clinical AI product with physicians and standalone frontier LLMs, each conducting an adaptive consultation and producing diagnostic and management outputs, therefore addresses an important evaluation question.

Language was also part of the evaluated condition. A Polish medical-examination benchmark containing 22,604 valid questions included paired Polish--English LEK and LDEK subsets and identified model- and examination-dependent cross-language differences,\cite{grzybowski2025} while a translation-controlled clinical-vignette study identified a model-by-language interaction, with some models performing worse in Polish and others showing similar overall performance across languages.\cite{ibrahimelnur2026} These findings motivate direct evaluation in Polish-language consultations.

Here, we conducted a controlled reader-by-case study using 150 synthetic Polish-language primary-care cases administered as adaptive simulated consultations. Doctorina was compared with eight physicians practising or training in family or internal medicine in Poland and with Google's Gemini 3.1 Pro Preview, Anthropic's Claude Opus 5, OpenAI's GPT-5.6-sol and Moonshot AI's Kimi K3. The analytical hierarchy designated the case-standardized difference between Doctorina and the physician cohort in Top-1 concordance with the designated primary diagnosis as the primary comparison. Primary-or-reference-differential concordance was the secondary diagnostic outcome; physician-reviewer ratings of diagnostic workup and initial treatment and comparisons with the standalone LLMs were exploratory. All groups were assessed on the same case corpus, clinical task and outcome definitions.

\section{Results}

\subsection{Analytical dataset}

All 150 synthetic Polish-language cases contributed to the diagnostic analyses. The diagnostic analysis included 240 assigned responses from eight physicians---five specialists and three residents: 60 cases had one physician response and 90 had two. Management outcomes were available for 239 responses, with at least one retained assessment for every case.

Both eligible Doctorina executions attempted all 150 cases. Run 481 yielded 148 technically valid diagnostic outputs and 147 technically valid management records from 149 recorded management attempts. Run 479 yielded 145 technically valid diagnostic outputs and 143 technically valid management outputs.

Gemini 3.1 Pro Preview, Claude Opus 5 and Kimi K3 yielded final outputs for all 150 cases, while GPT-5.6-sol yielded 148. The case-standardized six-group analyses comprised 150 equally weighted diagnostic cases and 149 complete management case pairs. Doctorina and physician outputs were assessed within one review block, while the four standalone-model outputs were assessed within a second aligned block using the same case references, endpoint definitions and rating direction. The six-group results combine a within-block Doctorina--physician comparison with aligned cross-block comparisons involving the standalone models.

\subsection{Diagnostic concordance}

\begin{figure}[!htbp]
\centering
\includegraphics[width=\textwidth]{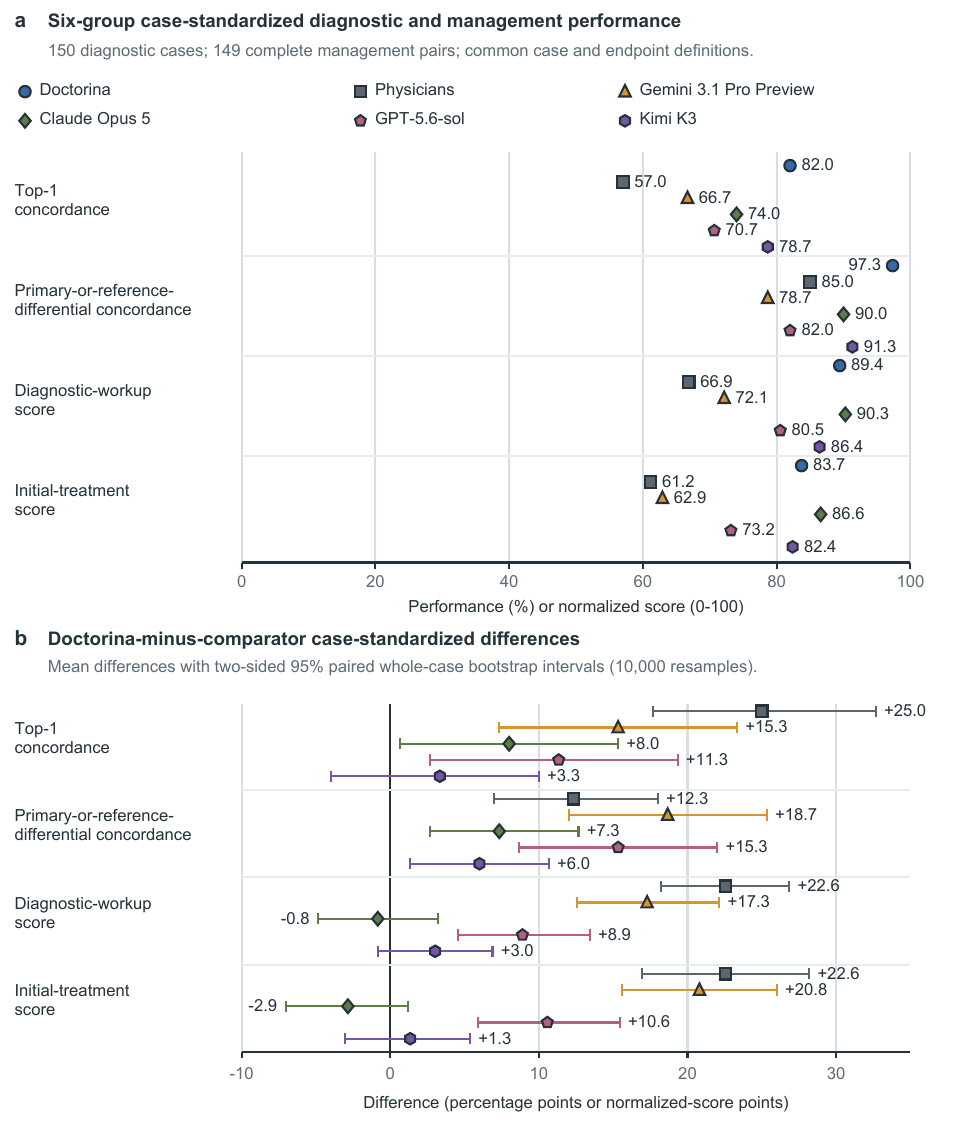}
\caption{Six-group case-standardized diagnostic and management performance. a, Top-1 concordance, primary-or-reference-differential concordance, diagnostic-workup score and initial-treatment score for Doctorina, physicians, Gemini 3.1 Pro Preview, Claude Opus 5, GPT-5.6-sol and Kimi K3. b, Paired Doctorina-minus-comparator differences; positive values favour Doctorina. Error bars show pointwise, two-sided 95\% percentile intervals from 10,000 paired whole-case bootstrap resamples. Doctorina was represented by run 481, physicians by within-case means and each standalone model by one output. Diagnostic analyses included 150 cases and management analyses 149 complete case pairs. Doctorina and physician outputs were reviewed together; standalone-model outputs were reviewed in a second aligned block. Only the Doctorina--physician Top-1 comparison was primary; the broader diagnostic endpoint was secondary, and management and standalone-model comparisons were exploratory. Diagnostic effects are reported in percentage points and management effects in normalized-score points.}\label{fig:performance}
\end{figure}

For the primary analytical comparison within the Doctorina--physician review block, concordance was 82.0\% for Doctorina and 57.0\% for physicians, a difference of 25.0 percentage points (95\% paired-case bootstrap CI, 17.7 to 32.7; Fig.~\ref{fig:performance}). In the aligned comparisons with the separately reviewed standalone-model outputs, Top-1 concordance was 78.7\% for Kimi, 74.0\% for Opus, 70.7\% for GPT and 66.7\% for Gemini. Doctorina-minus-model differences were 3.3 percentage points for Kimi, 8.0 points for Opus, 11.3 points for GPT and 15.3 points for Gemini; the respective 95\% CIs were -4.0 to 10.0, 0.7 to 15.3, 2.7 to 19.3 and 7.3 to 23.3.

The small aggregate separation between Doctorina and the two highest-scoring standalone models reflected predominantly concordant case-level classifications. Doctorina and Kimi had the same Top-1 status in 121 of 150 cases (80.7\%); Doctorina alone was concordant in 17 cases and Kimi alone in 12. Doctorina and Opus had the same status in 116 cases (77.3\%), with 23 Doctorina-only and 11 Opus-only concordant cases.

For the secondary primary-or-reference-differential endpoint, concordance was 97.3\% for Doctorina and 85.0\% for physicians, a difference of 12.3 percentage points (95\% CI, 7.0 to 18.0). Standalone-model concordance was 91.3\% for Kimi, 90.0\% for Opus, 82.0\% for GPT and 78.7\% for Gemini; the corresponding Doctorina-minus-model differences were 6.0, 7.3, 15.3 and 18.7 percentage points, with respective 95\% CIs of 1.3 to 10.7, 2.7 to 12.7, 8.7 to 22.0 and 12.0 to 25.3.

\subsection{Diagnostic-workup and initial-treatment outcomes}

On the exploratory 0--100 normalized management scale, diagnostic-workup scores across 149 complete case pairs were 90.3 for Opus, 89.4 for Doctorina, 86.4 for Kimi, 80.5 for GPT, 72.1 for Gemini and 66.9 for physicians (Fig.~\ref{fig:performance}). Doctorina-minus-comparator differences were 22.6 normalized-score points for physicians (95\% CI, 18.2 to 26.8), 17.3 points for Gemini (12.6 to 22.1), -0.8 points for Opus (-4.9 to 3.2), 8.9 points for GPT (4.5 to 13.4) and 3.0 points for Kimi (-0.8 to 6.9).

Initial-treatment scores were 86.6 for Opus, 83.7 for Doctorina, 82.4 for Kimi, 73.2 for GPT, 62.9 for Gemini and 61.2 for physicians. Doctorina-minus-comparator differences were 22.6 normalized-score points for physicians (95\% CI, 16.9 to 28.2), 20.8 points for Gemini (15.6 to 26.0), -2.9 points for Opus (-7.0 to 1.2), 10.6 points for GPT (5.9 to 15.4) and 1.3 points for Kimi (-3.0 to 5.4). Opus, Doctorina and Kimi had closely spaced management point estimates; the Doctorina-minus-Opus and Doctorina-minus-Kimi intervals spanned modest differences in either direction in both domains.

The case-weighted verdict distributions are shown in Table~\ref{tab:verdicts}. Doctorina, Opus and Kimi ratings were concentrated in categories 1 and 2 in both management domains, whereas physician ratings were distributed more broadly across the five categories. On the 150-case planned denominator, the case-weighted proportions receiving a poor or very poor verdict (categories 4--5) were 5.3\% for Doctorina, 20.7\% for physicians, 14.0\% for Gemini, 4.7\% for Opus, 6.7\% for GPT and 5.3\% for Kimi in workup. The corresponding treatment proportions were 7.3\%, 26.3\%, 20.7\%, 5.3\%, 10.7\% and 6.0\%, respectively.

\begin{table}[!htbp]
\caption{Physician-reviewer verdict distributions for diagnostic workup and initial treatment.}\label{tab:verdicts}
\fontsize{8.5}{10.5}\selectfont
\setlength{\tabcolsep}{2pt}
\renewcommand{\arraystretch}{1.12}
\begin{tabular}{@{}>{\raggedright\arraybackslash}p{16mm}>{\raggedright\arraybackslash}p{24mm}rrrrrrrrr@{}}
\toprule
\textbf{Domain} & \textbf{System} & \textbf{\shortstack{Numeric\\ratings}} & \textbf{\shortstack{1\\(\%)}} & \textbf{\shortstack{2\\(\%)}} & \textbf{\shortstack{3\\(\%)}} & \textbf{\shortstack{4\\(\%)}} & \textbf{\shortstack{5\\(\%)}} & \textbf{\shortstack{Invalid\\(\%)}} & \textbf{\shortstack{Unavailable\\(\%)}} & \textbf{\shortstack{Median\\(IQR)}} \\
\midrule
Diagnostic workup & Doctorina & 147 & 76.7 & 11.3 & 4.7 & 5.3 & 0.0 & 1.3 & 0.7 & 1 (1--1) \\
 & Physicians & 150 & 32.3 & 26.7 & 20.3 & 18.0 & 2.7 & 0.0 & 0.0 & 2 (1--3) \\
 & Gemini 3.1 Pro Preview & 150 & 40.0 & 26.7 & 19.3 & 10.0 & 4.0 & 0.0 & 0.0 & 2 (1--3) \\
 & Claude Opus 5 & 150 & 76.7 & 13.3 & 5.3 & 3.3 & 1.3 & 0.0 & 0.0 & 1 (1--1) \\
 & GPT-5.6-sol & 148 & 54.7 & 22.7 & 14.7 & 6.0 & 0.7 & 1.3 & 0.0 & 1 (1--2) \\
 & Kimi K3 & 150 & 64.7 & 22.0 & 8.0 & 4.7 & 0.7 & 0.0 & 0.0 & 1 (1--2) \\
\midrule
Initial treatment & Doctorina & 147 & 64.7 & 16.0 & 10.0 & 6.0 & 1.3 & 1.3 & 0.7 & 1 (1--2) \\
 & Physicians & 150 & 31.0 & 16.0 & 26.7 & 20.3 & 6.0 & 0.0 & 0.0 & 3 (1--4) \\
 & Gemini 3.1 Pro Preview & 150 & 31.3 & 21.3 & 26.7 & 9.3 & 11.3 & 0.0 & 0.0 & 2 (1--3) \\
 & Claude Opus 5 & 150 & 67.3 & 19.3 & 8.0 & 3.3 & 2.0 & 0.0 & 0.0 & 1 (1--2) \\
 & GPT-5.6-sol & 148 & 44.0 & 22.7 & 21.3 & 6.0 & 4.7 & 1.3 & 0.0 & 2 (1--3) \\
 & Kimi K3 & 150 & 58.7 & 20.0 & 15.3 & 4.7 & 1.3 & 0.0 & 0.0 & 1 (1--2) \\
\botrule
\end{tabular}
\footnotetext{Verdict columns are percentages of 150 planned cases. The physician category distribution therefore uses all 150 planned cases, whereas the physician management mean reported in Fig.~\ref{fig:performance} and the text is restricted to the 149 complete Doctorina--physician case pairs. Each case contributes equal weight; the physician rows summarize 239 response ratings across 150 cases, with each of two ratings contributing half of the case's weight when both were present. Medians and interquartile ranges (IQRs) use the same case weights. The scale was ordered from 1 (Excellent) to 5 (Very Poor). All-attempt normalized scores used the planned denominator, with 0 assigned to technical-error and incomplete-output states. The Invalid attempt column comprises Doctorina records labelled Technical error and GPT attempts without a final output; the Record unavailable column represents the single absent run-481 management record. Verdict-category summaries use observed numeric ratings. Doctorina and physician outputs were evaluated within one review block, and the four standalone-model outputs within a second aligned block.}
\end{table}

\subsection{Consistency across observed physician assignments}

Across the five disjoint physician packet-membership blocks, Doctorina-minus-physician differences were positive in every block for Top-1 concordance, diagnostic workup and treatment. Top-1 differences ranged from 10.0 to 46.7 percentage points; the corresponding ranges were 0.0 to 33.3 percentage points for primary-or-reference-differential concordance, 13.3 to 30.8 normalized-score points for workup and 14.7 to 38.3 points for treatment. The lower bound of 0.0 for the broader diagnostic endpoint represented one block-level tie. In respondent-specific paired comparisons, Doctorina-minus-physician Top-1 differences ranged from 6.7 to 46.7 percentage points across the eight physicians (Supplementary Fig.~1). Complete management-pair counts ranged from 28 to 30 per physician. No block- or respondent-specific point estimate favoured physicians, although ties occurred on the broader diagnostic endpoint and some intervals included zero in these 28--30-case comparisons.

\subsection{Exploratory case-level heterogeneity and diagnostic-error overlap}

Complexity levels 1, 2 and 3 contained 97, 38 and 15 diagnostic cases, respectively; the corresponding complete management strata contained 96, 38 and 15 cases. Doctorina Top-1 concordance across these levels was 83.5\%, 78.9\% and 80.0\%, compared with 63.9\%, 47.4\% and 36.7\% for physicians. Initial-treatment scores were 87.8, 80.9 and 65.0 for Doctorina and 66.8, 53.6 and 44.2 for physicians. Although the Doctorina--physician Top-1 point-estimate margin was larger at higher recorded complexity, all pairwise complexity-by-group interaction intervals for the physician comparison spanned zero across the four outcomes. The stratified estimates were therefore compatible with similar Doctorina--physician margins across the recorded complexity levels. Six-group estimates for every outcome and complexity stratum are reported in Supplementary Data 10.

Across 149 Doctorina cases with complete management data, Spearman correlations between Top-1 concordance and diagnostic-workup and treatment scores were 0.243 and 0.142, respectively. Across all six groups, the corresponding Top-1 correlations ranged from 0.073 to 0.267 for workup and from 0.093 to 0.289 for treatment. Twenty-three Doctorina cases combined Top-1 concordance with at least one management score below the Good threshold, whereas 14 combined Top-1 discordance with Good-or-Excellent scores in both management domains.

The five AI systems were all Top-1 concordant in 70 of 150 cases and all discordant in eight; exactly one AI system was concordant in nine cases. Doctorina was the sole concordant AI system in six cases. Of the 27 cases in which Doctorina's Top-1 diagnosis was discordant, at least one standalone LLM was concordant in 19. Of the eight cases in which all five AI systems were Top-1 discordant, seven were also Top-1 discordant for every observed physician response. These case-level patterns quantified diagnostic-concordance overlap among the five observed AI-system outputs and the physician cohort.

\subsection{Execution-level sensitivity and observed run agreement}

Run 479 reproduced positive Doctorina-minus-physician differences across all four outcomes: 23.7 percentage points for Top-1 concordance, 10.3 points for primary-or-reference-differential concordance, 18.3 normalized-score points for workup and 16.9 points for treatment; the respective 95\% CIs were 16.0 to 31.3, 4.7 to 16.0, 13.1 to 23.4 and 10.7 to 23.0.

The comparisons with the two highest-scoring standalone models were more sensitive to the eligible Doctorina execution. With run 479, Doctorina-minus-Opus differences were 6.7 percentage points for Top-1 concordance, 5.3 points for primary-or-reference-differential concordance, -4.8 normalized-score points for workup and -8.3 points for treatment; the respective 95\% CIs were -0.7 to 14.0, 0.0 to 10.7, -9.8 to 0.0 and -12.8 to -4.0. Doctorina-minus-Kimi differences were 2.0 percentage points for Top-1 concordance, 4.0 points for primary-or-reference-differential concordance, -1.0 normalized-score point for workup and -4.2 points for treatment; the respective 95\% CIs were -4.7 to 8.7, -1.3 to 9.3, -6.0 to 3.8 and -9.5 to 0.8. Diagnostic point estimates relative to Opus and Kimi remained positive in both executions, while management comparisons varied by execution.

Across all attempted cases from runs 479 and 481, Top-1 status agreed in 128 of 150 cases (85.3\%; Cohen's $\kappa$ = 0.517), and primary-or-reference-differential status agreed in 141 cases (94.0\%; $\kappa$ = 0.153). Among the 143 cases with technically valid diagnostic outputs from both executions, Top-1 status agreed in 126 cases (88.1\%; $\kappa$ = 0.567), and broader-endpoint status agreed in 141 (98.6\%; $\kappa$ = 0.493). Across all 149 matched management attempts, exact all-attempt normalized-score agreement was observed in 121 cases (81.2\%) for workup and 110 (73.8\%) for treatment. Among the 142 technically valid management pairs, exact verdict-category agreement was observed in 119 cases (83.8\%) for workup and 108 (76.1\%) for treatment. Workup and treatment were each within one verdict category in 133 cases (93.7\%); 124 cases (87.3\%) were within one category in both domains. Run 481 exceeded run 479 by 4.2 workup points (95\% CI, 0.5 to 8.4) and 5.5 treatment points (1.8 to 9.4) across 149 matched attempts; among 142 technically valid pairs, the differences were 0.9 points (-1.9 to 3.7) and 2.8 points (-0.2 to 5.8), respectively. Complete case-level robustness, heterogeneity, overlap and observed run-agreement tables are provided in Supplementary Data 10.

\section{Discussion}

Doctorina's principal advantage relative to physicians was higher performance across both diagnosis and management after adaptive consultation. Within the shared Doctorina--physician review block, the 25.0-percentage-point difference in Top-1 concordance was accompanied by higher workup and treatment scores and smaller proportions of Poor or Very Poor management ratings. This combination is clinically relevant to the evaluated task: stronger primary-diagnosis selection was accompanied by higher-rated investigation and initial treatment plans. Positive physician-comparison differences were reproduced in the second eligible execution and across the assignment-stratified point estimates, apart from ties on the broader diagnostic endpoint. In exploratory cross-block comparisons, Doctorina's point estimates exceeded Gemini and GPT across all four outcomes and led all standalone models on both diagnostic endpoints. Management estimates placed Opus, Doctorina and Kimi in a closely spaced leading group, with Opus numerically highest. Doctorina's comparative strength was therefore broad relative to the physician cohort and predominantly diagnostic relative to the strongest standalone models.

These comparisons are best interpreted in relation to the adaptive consultation task. Unlike complete-vignette evaluations, adaptive assessment makes performance contingent on respondent-directed information gathering. AMIE (Articulate Medical Intelligence Explorer) showed higher diagnostic accuracy than primary-care physicians in randomized, double-blind crossover text consultations with validated patient actors,\cite{tu2025} whereas general-purpose LLMs evaluated on cases derived from real patient data performed below physicians and deteriorated further when required to gather information stepwise.\cite{hager2024} In a randomized diagnostic-reasoning trial, an LLM performed well alone, while physician access to the LLM did not significantly improve diagnostic-reasoning scores compared with conventional resources.\cite{goh2024} Together, these studies show that comparative performance depends on the information environment, interaction design and role assigned to the AI. The current findings extend this evidence to an integrated clinical AI product evaluated in Polish through adaptive multi-turn consultations in which information acquisition preceded the scored diagnostic and management outputs.

Doctorina's product-level performance reflects the joint operation of its underlying language models and clinical scaffold; the present comparison does not isolate their respective contributions. General-purpose frontier models have surpassed specialized clinical tools on some benchmarks,\cite{vishwanath2026} while agent systems have achieved more modest gains at appreciably greater computational cost in others.\cite{liu2026agent} Replacing Doctorina's underlying models with models more capable in their assigned clinical tasks could improve system-level performance, but this remains an untested hypothesis. Controlled studies could vary the underlying models while holding the scaffold, case corpus and evaluation conditions constant, and complement these comparisons with component ablations under a fixed model configuration. Such studies should assess diagnostic and management quality alongside safety, technical completion and computational cost to determine whether model replacement improves the complete system.

The case-level findings further distinguished diagnostic prioritization from management quality. Most Doctorina Top-1 discordances still matched an accepted reference-differential diagnosis, indicating prioritization within the accepted set rather than complete diagnostic-set exclusion. The weak association between diagnostic concordance and management scores across groups further showed that diagnosis, workup and treatment capture distinct dimensions. Correspondingly, some Doctorina cases paired a correct primary diagnosis with sub-Good management, whereas others paired a discordant primary diagnosis with Good-or-Excellent management. These results support separate assessment and optimization of primary-diagnosis ranking, diagnostic workup and treatment.

Agreement across the two eligible executions provided a separate product-performance dimension. Diagnostic endpoints showed high raw agreement, although chance-corrected estimates were lower, particularly for the near-ceiling broader endpoint, and increased after technical failures were excluded. Management agreement was less exact. All-attempt differences favoured run 481, whereas valid-pair intervals included zero, indicating that technical completion contributed to the observed run-level difference; comparisons with Opus and Kimi likewise varied by execution. These findings support endpoint-specific agreement reporting and repeated-execution testing for closely ranked systems.

The complexity analysis refined the interpretation of the aggregate performance. Doctorina's Top-1 point estimates remained broadly similar across the three recorded complexity levels, whereas physician point estimates declined descriptively; Doctorina--physician point-estimate differences were positive in every stratum. Interaction intervals were compatible with no complexity-related change in the Doctorina--physician margin for each outcome, and the 15-case level-3 stratum produced wide intervals. Management point estimates were lower in the recorded higher-complexity strata for Doctorina and most comparator groups.

Diagnostic errors were partly shared and partly complementary across the five AI systems. Inter-system disagreement may therefore identify cases warranting additional review, whereas cases missed by every AI system and every observed physician response provide a focused set for examining reference ambiguity, incomplete elicitation, atypical presentation and common reasoning failure.

The Polish-language setting provides a locally relevant contribution. Prior Polish medical benchmarks have shown model- and task-dependent cross-language performance in examinations and translated vignettes.\cite{grzybowski2025,ibrahimelnur2026} This study adds adaptive history-taking and physician-rated management in the intended language and consultation format.

Several design features strengthen the evidence. All groups used one 150-case corpus, common constructed references, shared endpoint definitions and a common permitted-facts boundary. Case standardization gave each case equal weight regardless of whether one or two physician responses were available. The adaptive design embedded respondent-directed information seeking within the consultation that generated the diagnostic and management outputs. Whole-system analyses retained technical failures, while ordinal distributions separated rated content from failure burden. The central Doctorina--physician finding was examined in a second eligible execution and across physician, packet-membership and complexity analyses.

The controlled synthetic corpus enabled standardized comparison across a broad range of conditions, with the 150 cases distributed in proportion to the relative frequencies of selected eligible main ICD-10 diagnosis codes in 2024 National Health Fund service-reporting data. The designated diagnoses were established through the protocol's three-clinician review procedure and served as the constructed study reference for concordance assessment. The stochastic simulator produced respondent-specific histories within a common permitted-facts boundary. Within this design, the scored construct was final-output performance after adaptive text consultation. Simulator fidelity, information elicitation and consultation efficiency were outside the scored endpoints, and the evaluation did not encompass routine clinical workflow or patient outcomes. Generalization to clinical practice therefore requires prospective validation.

The physician comparison represented eight purposively recruited clinicians who completed partially crossed 30-case assignments in a single, approximately six-hour, resource-restricted text session. The analytical records did not preserve the packet-allocation rule, the number of physicians approached or reasons for non-participation, limiting assessment of recruitment and assignment effects. The purposive, non-randomized physician sample and reader-by-case structure constrain generalization to broader clinician populations, and the comparison estimates the complete Doctorina product relative to unaided physicians under the recorded study conditions. Doctorina--physician and standalone-model outputs were evaluated in separate, source-visible review blocks. The first block retained one final rating per output but not its achieved reviewer-pool size, output-level evaluator assignments, duplicate ratings or escalation identifiers. Consequently, cross-block AI contrasts may incorporate reviewer-pool and presentation effects, execution and scoring variation cannot be separated fully, and inter-rater reliability could not be estimated. A common blinded review pool, repeated ratings and broader physician sampling would strengthen the next validation.

The four Doctorina run identifiers referred to separate executions of the same case corpus. Under the development lead's branch-eligibility mapping, runs 478 and 480 were excluded, leaving runs 479 and 481 eligible for analysis. Run 481 was selected for the primary analysis because it contained more technically valid diagnostic and management records; run 479 was analysed separately as the execution-level sensitivity analysis. Because technical completeness contributed directly to all-attempt outcomes, this selection rule may favour the primary execution on those outcomes. Each standalone model contributed one corpus-wide execution, so cross-system estimates characterize the recorded execution samples. Routes, prompts, concurrency and completion states were available, whereas batch-wide generation settings and a run-linked manifest documenting the deployed Doctorina build and component configuration were unavailable, limiting exact technical replication. The standalone-model executions followed the physician--Doctorina comparison by more than two months; cross-system contrasts may therefore include temporal variation alongside system differences.

The normalized management score assumes equal spacing between adjacent ordinal categories; category distributions and medians therefore provide complementary summaries. The retained dataset provided a source-assigned complexity level for each case but did not include the corresponding classification criteria or assessor attribution; the stratified analyses were interpreted descriptively.

The primary analytical hierarchy was post-specified because the retained review field operationalized reference-differential concordance rather than ranked Top-3 accuracy. The retained protocol specified Top-1 and ranked Top-3 diagnostic accuracy as co-primary outcomes, each at a two-sided alpha of 0.025. This timing limits confirmatory interpretation. The Doctorina--physician Top-1 contrast was treated as primary; the remaining diagnostic, management, standalone-model, complexity and post hoc analyses were secondary or exploratory. The standalone-model analysis plan was finalized after output generation and before completion of the standalone-model physician-review block. Confidence intervals were pointwise without multiplicity adjustment. The analytical sample comprised the available 150-case corpus and eight completed physician assignments rather than a formal power target, and bootstrap intervals quantify case variation conditional on the observed physicians and executions. These conditions define the inferential scope of the reported estimates.

Prospective multicentre evaluation should assess the complete Doctorina system in routine clinical workflows, incorporating repeated executions, complexity-enriched sampling, disagreement-triggered review and patient-relevant outcomes while preserving distinct diagnostic and management endpoints.

\section{Methods}

\subsection{Study design and analytical scope}

This controlled reader-by-case study compared six respondent types in Polish-language simulated primary-care consultations: physicians, the Doctorina clinical AI system, Gemini 3.1 Pro Preview, Claude Opus 5, GPT-5.6-sol and Kimi K3. Eight physicians completed allocated subsets of the 150-case synthetic corpus. Two branch-correct Doctorina executions and one execution of each standalone large language model (LLM) attempted all 150 cases.

\Needspace{4\baselineskip}
Physician consultations and the two eligible Doctorina executions were conducted on 17 May 2026; the standalone-model consultations followed on 3--4 August 2026.

All groups used the same case corpus, constructed references, clinical task and outcome definitions. Case-standardized effects quantified the performance of each complete AI implementation---including its model, instructions, orchestration, state management and patient-facing presentation---or the observed physician cohort under this shared task. In every adaptive consultation, the final diagnostic and management outputs were contingent on the questions selected by the respondent. The case-based comparative design was informed by published vignette- and simulation-based evaluations of clinical AI systems and physicians.\cite{hager2024,tu2025,mcduff2025,gilbert2020,goh2024}

\begin{figure}[!htbp]
\centering
\includegraphics[width=\textwidth]{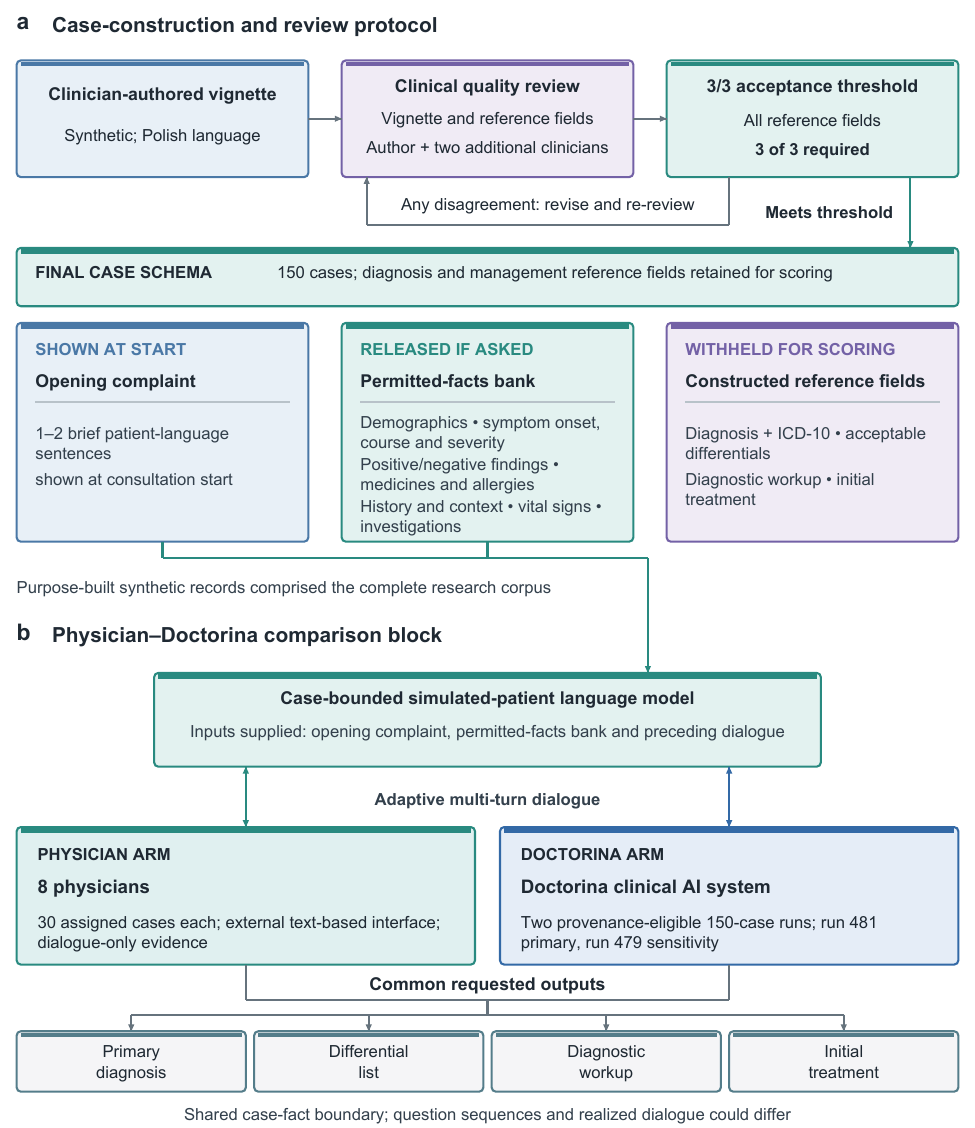}
\caption{Synthetic-case construction and physician--Doctorina consultation architecture. a, Each of the 150 synthetic Polish-language cases comprised three controlled information layers: an opening complaint shown at consultation start, a permitted-facts bank released through questioning and constructed diagnosis and management reference fields withheld for scoring; every field required approval by three clinicians. b, Physicians and Doctorina completed adaptive consultations through the same case-bounded simulated patient and returned a primary diagnosis, differential diagnoses, diagnostic workup and initial treatment. Eight physicians received 30 cases each; provenance-eligible runs 479 and 481 attempted all 150 cases.}\label{fig:architecture}
\end{figure}

Relevant items from STARD 2015, STARD-AI and TRIPOD-LLM informed reporting.\cite{bossuyt2015,sounderajah2025,gallifant2025} The study design and information boundaries are summarized in Fig.~\ref{fig:architecture}.

\subsection{Setting and physician participants}

The physician comparison was conducted during one approximately six-hour in-person session in Warsaw, Poland. Recruitment was purposive and used direct professional outreach. Eligibility required a medical degree, a valid licence to practise in Poland, current practice or postgraduate training in family medicine, general practice or internal medicine and availability for the complete session. Eligibility also required professional independence from Doctorina development, prompting, evaluation design and case creation, with first exposure to the analytical case set occurring during the assessed procedure.

Eight physicians completed the study: five specialists and three second-year residents. The study procedure provided each participant with access to the study interface and one non-analytical test case on the day before the assessed session to practise the interface and response format. During the assessed session, all eight physicians worked contemporaneously in the same controlled space under common instructions; an independent observer unaffiliated with Doctorina was present with responsibility for monitoring protocol adherence and recording any deviations or technical incidents. Physician participation and output review were compensated activities. Participants used numerical study identifiers and were distinct from the physicians who evaluated the outputs.

\subsection{Clinical-vignette corpus and physician allocation}

Clinician authors deliberately constructed the 150 synthetic cases (E1025--E1174) using a common template. The case authors were external to Doctorina and separate from the physician comparator, prompt-tuning and model-behaviour-optimization teams. The diagnostic composition of the corpus was informed by 2024 National Health Fund (NFZ) patient counts for services reported under main ICD-10 diagnosis codes at the five-character level. Eligible codes represented clinically informative underlying diagnoses reported for at least five patients; administrative encounter and symptom-only codes were outside the eligible diagnosis set. Relative frequencies were calculated across the eligible diagnoses; the most common eligible conditions were then selected and scaled to define the 150-case diagnostic distribution.\cite{nfz2024} Each vignette contained a presenting complaint of one or two short sentences and a separate permitted-facts bank covering demographics; symptom onset, course, character and severity; aggravating and relieving factors; relevant positive and negative findings; medicines, allergies and previous disease; family and social context; vital signs; and investigations available at presentation. Withheld reference fields specified the designated primary diagnosis and ICD-10 code, acceptable differential diagnoses, diagnostic workup, initial treatment and case-specific critical conditions.

The final case-review protocol required field-by-field review by the author and two additional clinicians, a three-of-three acceptance threshold and renewed review after revision of any disputed field. The accepted final fields constituted the constructed case references. The corpus contained 127 distinct recorded disease labels and 128 distinct ICD-10 values; 82 cases described female patients and 68 described male patients. The age distribution is summarized in Table~\ref{tab:ages}. The source case table recorded complexity on an ordinal scale from level 1 (lowest complexity) to level 3 (highest complexity): 97 cases were level 1, 38 were level 2 and 15 were level 3; these source classifications defined the exploratory strata. A named comorbidity was recorded in 119 cases.

\begin{table}[!htbp]
\caption{Age distribution of the synthetic primary-care case corpus.}\label{tab:ages}
\setlength{\tabcolsep}{12pt}
\renewcommand{\arraystretch}{1.12}
\begin{tabular}{@{}lrr@{}}
\toprule
\TCH{Age group, years} & \TCH{Cases, n} & \TCH{\%} \\
\midrule
0--1 & 2 & 1.3 \\
2--5 & 3 & 2.0 \\
6--12 & 3 & 2.0 \\
13--17 & 8 & 5.3 \\
18--25 & 14 & 9.3 \\
26--39 & 39 & 26.0 \\
40--59 & 40 & 26.7 \\
60--74 & 32 & 21.3 \\
75 or older & 9 & 6.0 \\
Total & 150 & 100.0 \\
\botrule
\end{tabular}
\footnotetext{Note: Counts and percentages are shown separately. Percentages use the full 150-case denominator and may not sum to 100\% because of rounding.}
\end{table}

Ten packets of 30 cases formed five paired case-membership sets. The two packets in each pair contained the same cases in different orders, placing every case in two packet numbers. One packet was assigned to each physician. Eight completed assignments yielded 240 retained physician diagnostic consultations: 90 cases had two physician responses and 60 had one.

\subsection{Consultation environment and physician procedure}

Cases were administered as adaptive text dialogues. Each dialogue opened with the presenting complaint. Matched execution traces identified o3-2025-04-16 as the simulated-patient model; the simulator template specified a temperature of 1, and the model generated replies from the case-specific permitted-facts bank and preceding dialogue. The simulator was instructed to answer briefly and truthfully in the language of the case, remain within case-supported facts, preserve diagnostic neutrality and state when the case information did not resolve a request. Messages containing several unrelated questions elicited responses to the first one or two and could prompt brief clarification. Consultations used a fixed evidence environment comprising the investigations documented in each case. These rules standardized the factual boundary while allowing each respondent to determine the sequence and content of questioning.

Physicians completed each consultation independently through an external text-based conversational interface and determined when to end the exchange. Each started case was completed before submission. The assessment interface presented the case dialogue as the sole evidence source, and physicians worked independently under standardized closed-resource conditions. At completion, each physician submitted a most-likely diagnosis, differential diagnoses, recommended diagnostic workup and initial treatment or medication plan. Diagnostic evaluation used semantic concordance of the submitted free-text diagnosis; ICD-10 coding was optional.

\subsection{AI systems and executions}

The study protocol identified Doctorina as version 2.1.3 and described an integrated clinical AI product comprising doctor, assistant, final-recommendation, translation, attachment-processing, reference and follow-up functions coordinated by a routing agent. In the \href{https://github.com/DoctorinaAI/doctorina/blob/5f134f472f9700af9b72d2b2b813ca0023a33c52/backend/ai-doctor-core/src/aidoctor/core/doctorina/agents_factory.py}{source code for the protocol-linked v2.1.3 release} (commit 5f134f4), the configured OpenAI model identifiers were o3 for the doctor agent; \mbox{gpt-5-mini-2025-08-07} for the assistant, translation and follow-up agents; \mbox{gpt-5-2025-08-07} for the final-recommendation and attachment-processing agents; and \mbox{o3-mini-2025-01-31} for the references agent. The routing component dispatched between agents without a separately assigned language model.

Doctorina was evaluated as the complete product with the protocol-described native functions. The execution protocol used an isolated session for each case, presented the Polish opening complaint, routed questioning through the case-bounded simulator and required the same four clinical domains as the physician procedure. A fixed execution protocol was applied across cases, and the generated patient-visible outputs passed directly to evaluation.

Runs 478, 479, 480 and 481 were four separately recorded Doctorina executions, each attempting the same 150 cases; the identifiers denoted execution batches, not different case sets. Eligibility and analytical role were determined in two steps. First, the development lead's run-to-branch mapping identified runs 479 and 481 as originating from the software branch designated for the study. The same mapping placed runs 478 and 480 on a different branch, and these runs were excluded from the reported comparisons. Second, technical completeness determined the primary and sensitivity roles within the eligible pair.

For runs 479 and 481, case-level quality control reconciled the retained records against the 150-case corpus and reviewed logged completion and technical-error states. Run 481 yielded 295 technically valid records across the two retained domains---148 diagnostic outputs and 147 management records from 149 recorded management attempts---and supplied the primary Doctorina estimates. Run 479 yielded 288 technically valid records---145 diagnostic outputs and 143 management outputs---and was analysed separately for execution-level sensitivity. The patient-visible final recommendation from each eligible execution constituted the evaluated output. Full execution-level completion states are provided in Supplementary Data 10.

The standalone comparators were Gemini 3.1 Pro Preview, Claude Opus 5, GPT-5.6-sol and Kimi K3, executed through the recorded routes \path{vertex_ai/gemini-3.1-pro-preview}, \path{bedrock/eu.anthropic.claude-opus-5}, \path{openai/gpt-5.6-sol} and \path{moonshot/kimi-k3}, respectively. Gemini, Opus and GPT were executed on 3 August 2026, and Kimi on 4 August 2026. Following initial specification of Gemini, Opus and GPT, the final comparator set incorporated Kimi before completion of the standalone-model physician-review block. Each model underwent one corpus-wide execution comprising 150 attempted cases under the identical instruction, ``Pretend you're a doctor and consult a patient.'' The execution protocol supplied the same case-specific Polish opening complaint, used the same case-bounded simulated-patient design and required the same four clinical domains, with constructed reference fields reserved for evaluation. The concurrency limit was 150 conversations for Gemini, Opus and GPT and 20 for Kimi.

Each case began with its case-specific opening complaint and used a separate conversation history; the logs preserved the accumulated patient-visible dialogue within each case. The execution harness recorded conversation completion according to its chat-state classification. Gemini, Opus and Kimi yielded final outputs for all 150 cases. GPT yielded 148 final outputs; its two remaining attempts reached the 30-pair execution limit and were retained with outcome value 0 in all-attempt analyses. Kimi's five first-attempt provider rate-limit events were successfully retried within the same corpus execution. The permitted-facts bank and constructed reference fields remained within the simulated-patient and evaluation layers.

\subsection{Data capture and quality control}

Analyses integrated the case-distribution and packet workbooks, diagnostic and management-rating worksheets, model execution logs and the source case table. A case--respondent ledger linked source values and formulas to identifiers, provenance, completion states, inclusion flags and analytical outcomes. Analyses used Python 3.12.13, NumPy 2.3.5, pandas 2.2.3 and openpyxl 3.1.5.

Physician assignments were reconstructed by matching submitted cases to the five packet-pair sets. Matching produced a retained diagnostic cohort of 240 assigned observations; one additional E1164 response did not correspond to its respondent's reconstructed assignment and therefore remained outside the analytical cohort. The management cohort comprised 239 observed records spanning all 150 cases, with E1129 contributing to the diagnostic analysis.

Standalone-model datasets were validated for unique case ordinals, complete start and finish markers, opening-complaint agreement and completion-status consistency.

\subsection{Physician review and outcomes}

Physician and Doctorina outputs comprised one review block and underwent manual evaluation through a resident-physician review process distinct from the eight physician respondents. The review protocol specified one primary evaluator per output, with escalation of uncertain judgments to a senior physician.

Standalone-LLM outputs were evaluated subsequently in a second aligned review block by four physician reviewers using the same case references, endpoint definitions and rating direction. Cases were the allocation unit: the same reviewer evaluated all four standalone-model outputs for a given case. The realized allocation was 104, 26, 15 and 5 cases across the four reviewers. In both blocks, reviewers assessed the retained final outputs against the constructed case references. One final diagnostic judgment, workup verdict and treatment verdict was retained per evaluated output.

\subsubsection{Diagnostic concordance}

Top-1 concordance indicated semantic agreement between the submitted most-likely diagnosis and the designated primary diagnosis. Standard abbreviations, clinically equivalent terminology, synonyms and Polish--English equivalents were accepted when they denoted the same condition. For Doctorina, the diagnosis explicitly identified as primary or most likely in the patient-visible terminal recommendation defined the submitted primary diagnosis; for each standalone LLM, the first diagnosis listed in the retained final model output did so. Diagnosis text, rather than a submitted ICD-10 code, governed concordance for all groups.

After a Top-1 mismatch, a secondary binary field recorded whether the submitted primary diagnosis matched a condition in the designated reference differential. Primary-or-reference-differential concordance equalled 1 when either field equalled 1. This endpoint measured whether the submitted primary diagnosis belonged to the constructed primary-and-differential reference set.

\Needspace{14\baselineskip}
\subsubsection{Diagnostic workup and treatment}

Diagnostic workup and initial treatment were rated independently of diagnostic correctness on an ordered five-category scale: 1, Excellent; 2, Good; 3, Adequate; 4, Poor; and 5, Very Poor, with lower values representing better performance.

The review guideline anchored Adequate (3) to essential investigations or basic therapy addressing the patient's main needs, with further refinement possible. Poor (4) denoted important omissions, inappropriate investigations or treatment of questionable appropriateness; Very Poor (5) denoted critical under-investigation or a treatment plan judged highly likely to worsen the patient's condition. Categories 1--3 therefore represent Adequate-or-better management quality, and categories 4--5 performance below that level.

\Needspace{5\baselineskip}
For n numeric verdicts, the arithmetic mean verdict was transformed to a normalized descriptive score:

\[\text{Normalized score}=100\times\frac{5-\text{mean verdict}}{4}.\]

Mean verdicts 1--5 therefore mapped to 100, 75, 50, 25 and 0. Category distributions and medians with interquartile ranges were calculated among numeric physician-reviewer verdicts using the case weights defined below. The normalized score was used for case-standardized group comparisons.

\subsection{Technical completeness and analysis populations}

A technical error denoted a source-recorded failure state, labelled Error, NONE (Error) or Technical error, for which the affected outcome was classified as technically invalid in the analysis. These states were distinguished from clinically discordant diagnoses, numeric management ratings, consultations that reached the execution limit without a final output and administratively absent records.

All-attempt analyses preserved the planned denominators by assigning outcome value 0 to technical-error and incomplete attempts: diagnostic outcomes were coded 0 for both binary endpoints, and management outcomes received a normalized score of 0. Numeric verdicts were used for category distributions and medians, observed analytical records for available-record means and planned denominators for separately tabulated administrative absences.

\subsection{Statistics and reproducibility}

The final analytical hierarchy designated the Doctorina--physician Top-1 comparison as primary, primary-or-reference-differential concordance as secondary, and management and standalone-model comparisons as exploratory. The standalone-model analysis plan was finalized before completion of the standalone-model physician-review block.

The primary analytical comparison was the paired, case-standardized difference between Doctorina run 481 and the physician within-case mean in Top-1 concordance. The analytical sample comprised the complete accepted corpus of 150 cases and all eight completed physician assignments. Each diagnostic comparison included all 150 cases, and management comparisons comprised 149 complete case pairs; E1154 contributed a diagnostic record but not a management record in run 481. Doctorina contributed the run-481 value for each available case, physicians contributed the mean of one or two retained responses and each standalone LLM contributed one output. The Doctorina-minus-comparator effect was the mean paired within-case difference, reported in percentage points for diagnostic outcomes and normalized-score points for management outcomes.

Two-sided 95\% percentile confidence intervals were estimated from 10,000 paired whole-case bootstrap resamples with seed 20260730. Each resampled case carried all retained physician and system records, preserving the observed reader-by-case and review-block structure. The intervals quantified case-level variation conditional on the observed physician sample, system executions and retained final ratings. Confidence intervals were pointwise, and comparator and exploratory analyses were interpreted descriptively.

Workup and treatment category distributions were case weighted so that every rated case contributed one unit of probability mass. For cases with two physician ratings, each rating contributed one-half of that case's mass; single-rating cases contributed one complete unit. Doctorina run 481 and each standalone LLM contributed one case-level physician-reviewer verdict when available.

\begin{figure}[!htbp]
\centering
\includegraphics[width=\textwidth]{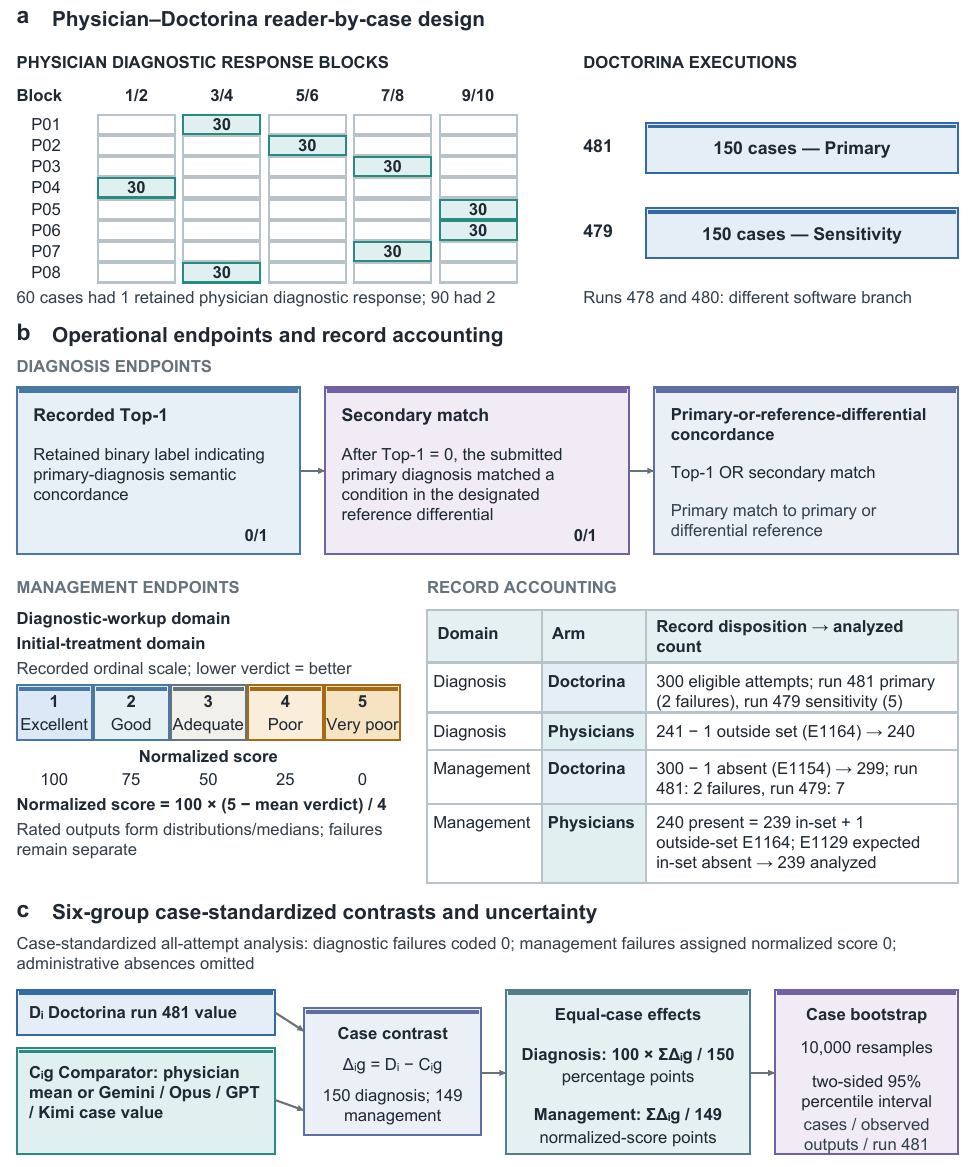}
\caption{Reader-by-case observation structure and case-standardized estimand. a, Run 481 served as the primary Doctorina execution and run 479 as the execution-level sensitivity analysis. The physician cohort contributed 240 diagnostic observations; 90 cases had two responses and 60 had one. b, Diagnostic endpoints compared the submitted primary diagnosis with the constructed reference diagnosis set; workup and treatment were rated on the five-category scale and transformed to normalized scores. c, Each case contributed equally to Doctorina-minus-comparator contrasts, with physicians represented by their within-case mean. Uncertainty was estimated using 10,000 paired whole-case bootstrap resamples.}\label{fig:estimand}
\end{figure}

Execution-level sensitivity analysis replaced primary run 481 with the second eligible execution, run 479. All 150 cases contributed to each run-479 sensitivity outcome. Fig.~\ref{fig:estimand} summarizes the observation structure, endpoints and case-standardized estimand.

Additional exploratory analyses characterized case-level robustness and heterogeneity. Poor or very poor management verdicts were defined as categories 4 or 5 and summarized on the 150-case denominator using the case weights and outcome categories defined above. Doctorina-centred win--tie--loss summaries classified each standardized case value as higher than, equal to or lower than the corresponding comparator value. Assignment-stratified consistency was examined by comparing run 481 separately with each physician across that physician's assigned cases and within the five identifiable packet-pair membership blocks. Complexity-stratified estimates used the source-recorded levels 1, 2 and 3; bootstrap resampling was performed within each stratum, and pairwise differences between stratum-specific Doctorina--comparator effects were summarized as exploratory interaction contrasts. Diagnostic-error overlap was summarized at case level across the binary outcomes of Doctorina, Gemini, Opus, GPT and Kimi and was cross-referenced with the observed physician responses on cases missed by every AI system. The relation between diagnostic correctness and management performance was summarized using Spearman rank correlations between binary diagnostic outcomes and normalized management scores. Case-level cross-classification used Good-or-Excellent management scores of at least 75 under the all-attempt coding defined above.

Observed agreement between the all-attempt records from runs 479 and 481 was summarized by raw agreement and Cohen's $\kappa$ for the binary diagnostic outcomes. Management agreement across all common attempts was summarized by exact agreement of the all-attempt normalized outcomes and paired mean differences. Within the technically valid subset, agreement was summarized by exact and within-one-verdict-category agreement and paired mean differences. These measures characterize agreement in the combined execution-and-evaluation record represented by the final retained ratings. Exploratory comparisons used the same bootstrap procedure. Reproducible case-level tables and supplementary analyses are provided in Supplementary Data 10.

\FloatBarrier\backmatter
\section*{Data availability}
The deidentified analytical data accompanying this article are provided as Supplementary Data 1--10. Supplementary Data 1--4 contain case metadata, physician allocation and coded Doctorina--physician and standalone-model rating ledgers; Supplementary Data 5--9 contain the case-standardized analysis ledger, eligible-execution sensitivity results, arm estimates, contrasts and verdict distributions; and Supplementary Data 10 contains validated robustness, heterogeneity, overlap and observed run-agreement tables. Deidentified source data for Fig.~\ref{fig:performance} and Supplementary Fig.~1 are included with the figure package. Researchers seeking to reproduce or verify the analyses may request access to retained case narratives, dialogue outputs and execution records from the corresponding author. Access is limited to the records required for the stated verification purpose and is subject to a signed confidentiality and data-use agreement and applicable data restrictions.

\section*{Code availability}
Supplementary Code 1 reproduces the primary six-group arm estimates and Doctorina-centred paired-bootstrap contrasts. Supplementary Code 2 validates the public data files, workbook structure and file hashes. Reproducible generation code for Supplementary Fig.~1 and validation code for Fig.~\ref{fig:performance} and Table~\ref{tab:verdicts} are included with the manuscript source package.

\section*{Acknowledgements}
The authors thank the physicians who authored and reviewed the synthetic clinical cases, the eight participating physicians, the independent session observer and the physician reviewers who evaluated the study outputs. The authors also thank Shubhanan Upadhyay for independent scientific and methodological advice and critical review of the manuscript. This study received no external grant funding.

\Needspace{12\baselineskip}
\section*{Author contributions}
H.P.: Conceptualization, methodology, resources, supervision and project administration. S.S.: Investigation, data curation, project administration, resources and formal analysis. A.K.: Software, data curation, formal analysis and methodology. A.N.: Formal analysis, visualization, writing---original draft, and writing---review and editing. P.G.: Methodology, validation, investigation, resources, project administration and supervision. J.M.: Methodology, validation, investigation, data curation and formal analysis. V.H.: Methodology, validation, investigation and data curation. A.R.: Software and resources. P.S.: Software and resources.

\section*{Competing interests}
H.P. is a co-founder and Chief Medical Officer of Doctorina. S.S. is Clinical Lead at Doctorina. A.K. contributed to the technical development of Doctorina and the study infrastructure. A.N. is an affiliated medical expert at Doctorina. P.G., J.M. and V.H. declare no competing interests. A.R. and P.S. contributed to the technical development of Doctorina and the study infrastructure.

\clearpage
\setlength{\bibsep}{5pt}

\end{document}